\documentclass{article}
\usepackage{amsmath,graphicx}
\usepackage[preprint]{spconf}
\usepackage{tabularx}
\usepackage{xcolor}
\usepackage{subcaption}
\usepackage[font={scriptsize}]{subcaption}

\copyrightnotice{\copyright\ IEEE 2018}
\toappear{{\it Proc.\ ICIP2018, Oct 7-10, 2018, Athens, Greece}}

\newenvironment{packed_enum}{
\begin{enumerate}
  \setlength{\itemsep}{2pt}
  \setlength{\parskip}{0pt}
  \setlength{\parsep}{0pt}
}{\end{enumerate}}

\title{Representing Images in 200 Bytes: Compression via Triangulation}
\name{David Marwood, Pascal Massimino, Michele Covell, Shumeet Baluja}
\address{Google, Inc}
\begin{document}
\maketitle
\begin{abstract}

  A rapidly increasing portion of internet traffic is dominated by
  requests from mobile devices with limited and metered
  bandwidth constraints.  To satisfy these requests, it has become
  standard practice for websites to transmit small and extremely
  compressed image previews as part of the initial page
  load process to improve responsiveness. Increasing thumbnail
  compression beyond the capabilities of existing codecs is therefore
  an active research direction.  In this work, we concentrate on
  extreme compression rates, where the size of the image is typically
  200 bytes or less.  First, we propose a novel approach for image compression
  that, unlike commonly used methods, does not rely on block-based
  statistics.  We use an approach based on an adaptive triangulation
  of the target image, devoting more triangles to high entropy regions
  of the image. Second, we present a novel algorithm for encoding the
  triangles.  The results show
  favorable statistics, in terms of PSNR and SSIM, over both the JPEG
  and the WebP standards.

\end{abstract}
\section{Introduction}
\label{sec:intro}

The need for highly compressed images continues to increase.  A
rapidly increasing portion of internet traffic is dominated by
requests from mobile devices with limited and often metered bandwidth
constraints.  Efficient delivery of quality thumbnails is an active
area of interest to some of the largest Internet companies, including
Google~\cite{webp}, Facebook~\cite{cabral2015} and Apple~\cite{apple}.
In addition to decreased download latency and bandwidth for end users,
reducing image size also helps with storage requirements for billions
of thumbnails that need to be rapidly accessed.

JPEG has long been a standard approach for image compression.  In this
study, we examine compression in an operating regime where JPEG and
other popular approaches do not fare well: under 200 bytes.
Usually, when extreme compression is required, it is addressed with
domain specific techniques, specialized for faces~\cite{bryt2008compression},
satellite imagery~\cite{huang2011satellite}, smooth synthetic
images~\cite{orzan2013diffusion}, and
surveillance~\cite{zhu2015dictionary}, among others.

A powerful, recent, image compression approach is \emph{WebP}.
Per~\cite{webp}, WebP lossless images are 26\% smaller in size
compared to PNGs. WebP lossy images are 25-34\% smaller than
comparable JPEG images at an equivalent SSIM quality index. This is
the standard to which we will compare. 

 The basis of many of the popular compression techniques is a
 subdivision of the image into a set of blocks.  Our approach, which
 is based on triangulation, does not use a block approach nor a
 predefined, or uniform, spacing of triangles over the image. Instead,
 we use a limited set of vertices which are assigned a color index
 from a small colormap; simple color interpolation between each of the
 triangle vertices is used to fill in the triangles to create the
 resultant image.  More triangles are devoted to the complex (high
 entropy) regions.  Triangulation has previously been used in a
 diverse set of approaches for compression, see
 ~\cite{bougleux2009image,davoine1996fractal,demaret2006image}.
 Finally, other experimental compression approaches use deep neural
 networks~\cite{jiang1999image,toderici2015} and
 diffusion~\cite{schmaltz2009,hoeltgen2018}.

\section{Triangulation of Images}
\label{sec:algorithm}

There are two broad components of our approach. The first is creating
an effective triangulation and the second is efficiently encoding the
triangulation.  The triangulation component can be thought of as two
pieces that must interact well: selecting where to place the
triangulation's vertices and assigning a single color to each vertex.
For transmission efficiency, we would like to minimize the number of
vertices and the total number of unique colors.  Rather than
transmitting the connectivity matrix, we consider only Delaunay
Triangulations that can be constructed in both the encoding and
decoding stages given only the vertex coordinates. See also~\cite{galperin1983succinct,turan1984succinct}.

Our approach follows a generate-and-test paradigm.  We begin with a
small thumbnail image, $I$
(usually $221 \times 221$).
In the simplest version of our algorithm, shown in
Figure~\ref{baseline}, we begin with an over-complete set of vertices
on a fixed-size grid and prune them to a smaller set until the set and
the color information can fit in 200 bytes. The grid-size is our only
parameter for adjusting the compression rate.

\begin{figure}[t!]  
  \noindent\fbox{\parbox{\dimexpr\linewidth-2\fboxsep-2\fboxrule\relax}{
    \begin{packed_enum}
      \small
      
 \item \emph{Insert initial, evenly-spaced candidate coordinate points
   from a subsampled version of $I$, into a set $P$.}
   By considering only a fixed subset $P$ of
   pixels locations for triangle coordinates, the encoding is faster
   and more compressible.

 \item \emph{Select a set $C$ of colors.}  Cluster the colors in $I$
   and select the 8 to 16 representative ones.  Each vertex $p \in P$ is
   assigned a color in $C$.

 \item   \emph{Triangulate.} Given $P$, create a Delaunay Triangulation.

 \item  \emph{Generate $I'$ and test.} With $C$ and $P$, 
  the triangles are filled
  using bilinear interpolation, yielding image  $I'$. The difference between $I$ and $I'$ is tested with either SSIM or PSNR.

 \item  \emph{Find least important vertex $p_x \in P$. } For each $p
  \in P$, in turn, remove that $p$, re-triangulate and fill triangles to create $I''$.  Set $p_x$ to the point that produces the lowest
  error.\label{repeat}

 \item  \emph {Remove least important vertex, $p_x$.}
  Continue from \ref{repeat}.
    \end{packed_enum}
      \vskip -0.05in    
  }}
      \vskip -0.05in  
\caption{Baseline: A Deterministic, Greedy, Approach.}
\label {baseline}
\vskip -0.1in
\end{figure}

\normalsize

\begin{figure}[b!]
\centering
  \includegraphics[width=0.85\linewidth]{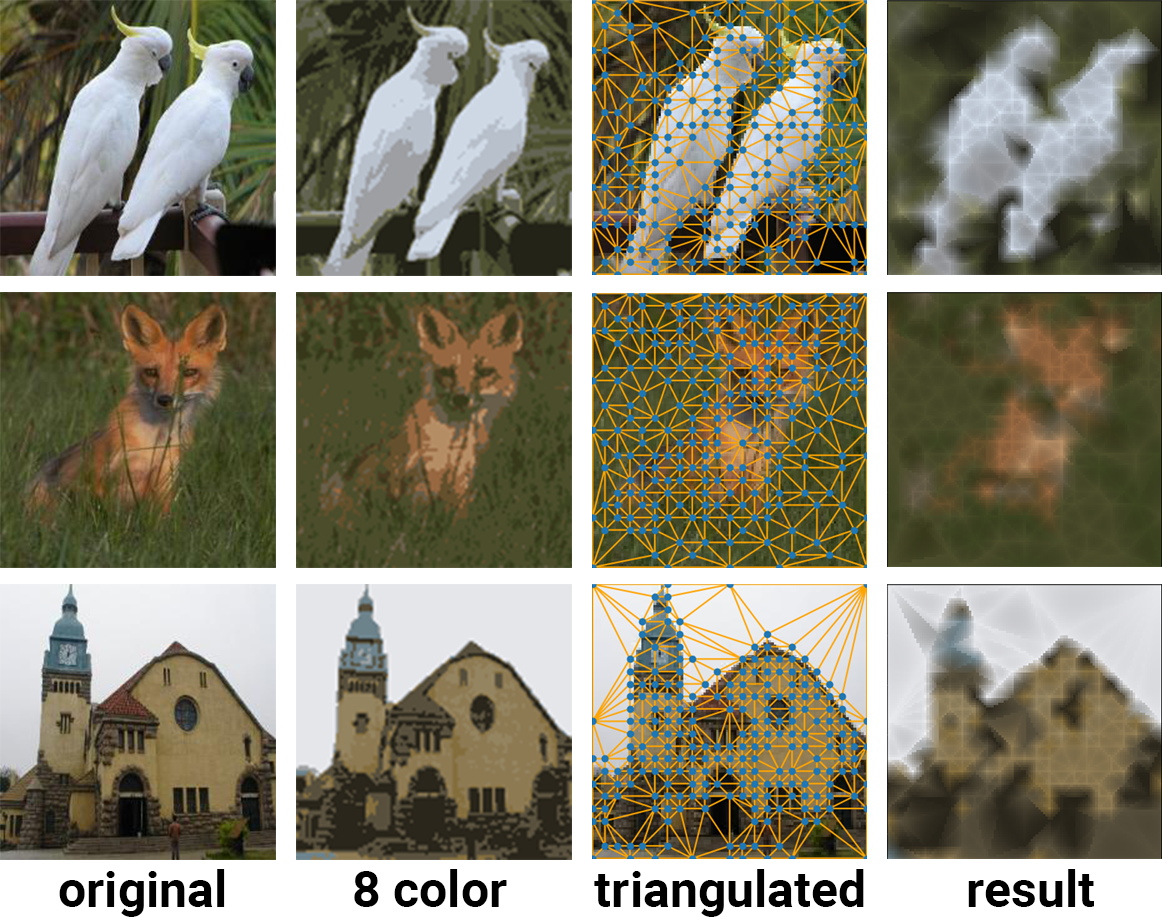}
\caption{Results for baseline.  From left: Original Image, 8 colors version, triangulation, and result.  Note that the triangulation places more triangles in higher entropy regions.}
  \label{baselineResults}
\end{figure}

Using only this simple procedure, the final triangulations and
resulting images are shown in Figure~\ref{baselineResults}.  The most
salient observation is that the triangles cluster around the higher
entropy regions.  Homogeneous regions such as the sky have fewer
triangles.  Quantitatively, PSNR quality was close to WebP, but did
not consistently out-perform WebP.  Understanding the deficiencies
is key to understanding the design of our improved approach.
(1) The colors, based on global image characteristics, are selected
once and never adapted.
(2) The only allowed change is to remove vertices: they are never added back
or moved slightly to find a better combinations of triangles.
(3) The Baseline makes a greedy choice for every proposed mutation,
limiting the effectiveness of local search once regions of high
performance are found.
  
To address the difficulties, we turn to a stochastic variant.  An
often-used technique to search discrete, non-differentiable,
optimization landscapes is with randomized search heuristics such as
hill climbing~\cite{muhlenbein1992genetic}, evolutionary algorithms and
strategies~\cite{dasgupta2013evolutionary,salimans2017} and simulated
annealing~\cite{kirkpatrick1983optimization}.

A stochastic hill climbing variant of the Baseline is given in
Figure~\ref{stochastic}.  At the expense of determinism, this
allows us to more thoroughly explore the search
space and adapt color settings.  There are six possible operations
other than the single ``remove vertex'' operator that the Baseline
variant employed. Note that 4 of 7 mutation operators (Steps 5(d)-(g) in
Figure~\ref{stochastic}) modify colors; no color modification was
employed in the Baseline.  Second, vertices can be added (even if they
had been previously removed) if they are found to improve the score;
vertices can also make ``localized'' moves to nearby grid points.
Third, instead of always removing vertices until the desired byte-size
was reached, the acceptance of a move is based on whether that move
improves the quality-vs-size objective function.

In contrast to the Baseline, the stochastic variant is initialized
with only a small set of vertices $P$.  As before, they are
all on a pre-specified grid.  The search progresses by \emph {mutating}
the current solution set and evaluating the result with respect to the
quality-vs-size objective function.

\begin{figure}[t!]
\noindent\fbox{\parbox{\dimexpr\linewidth-2\fboxsep-2\fboxrule\relax}{
\begin {packed_enum}
\small

\item \emph{Initialize $P$.} Greedily select $P$
  as a set of 300 vertices that minimizes the difference with the input
  image.

\item \emph{Initialize $C$.} Agglomerate
  the individual vertex colors
  down to 8 colors in the color table $C$.

\item \emph{Initialize vertex colors.} For each vertex, assign the
  color index in $C$ closest to the vertex's color in the input image.
  
\item \emph{Triangulate.} Given $P$, create a Delaunay triangulation.

\item \emph{Mutate.} A mutation is a subset of the actions
  below. An action is included (once) with some probability.   
  \begin {packed_enum}
  \item \emph{Displace a vertex.} Move a vertex either horizontally
    or vertically one grid point.
  \item \emph{Add a random vertex.}
  \item \emph{Remove a random vertex.}    
  \item \emph{Re-assign vertex color randomly.}
  \item \emph{Add a color to $C$ and re-assign vertex colors.}
  \item \emph{Remove a color from $C$ and re-assign vertex colors.}
  \item \emph{Perturb a color entry.} Select a color entry
    and a channel and randomly change by ${\pm 1}$. 

  \end {packed_enum}
\item \emph{Re-triangulate and retain the mutation if it improved the trade-off between error
  and size.  Otherwise discard the mutation.} Repeat from Step 5.
\end {packed_enum}
      \vskip -0.05in  
}}
      \vskip -0.05in    
\caption{A Stochastic Approach. }
\label {stochastic}
\vskip -0.1in
\end{figure}

\begin{figure*}[t]
  
  \centering

  \begin{minipage}{1.2in}
    \caption{\small
 Visualizing the results vs. image byte size.  For each of the
  3 original images (left most column), the image at 2 WebP (roughly 400, 100
  bytes) and 5 compression levels for our system (roughly 400, 300, 250, 200, 100)
  are given.  Byte size (b) and grid size (g) given for each result
  from our system.  }
\label{mainimages}
  \end{minipage}~~~~~
  \begin{minipage}{5.5in}
      ~~~original~~~~~~~~~~~~~~~~~~~~~WebP ~~~~~~~~~~~~~~~~~~~~~~~~~~~~~~~~~~~~~~~~~~~~~~~~~~~~~~ our system    
    
    \input {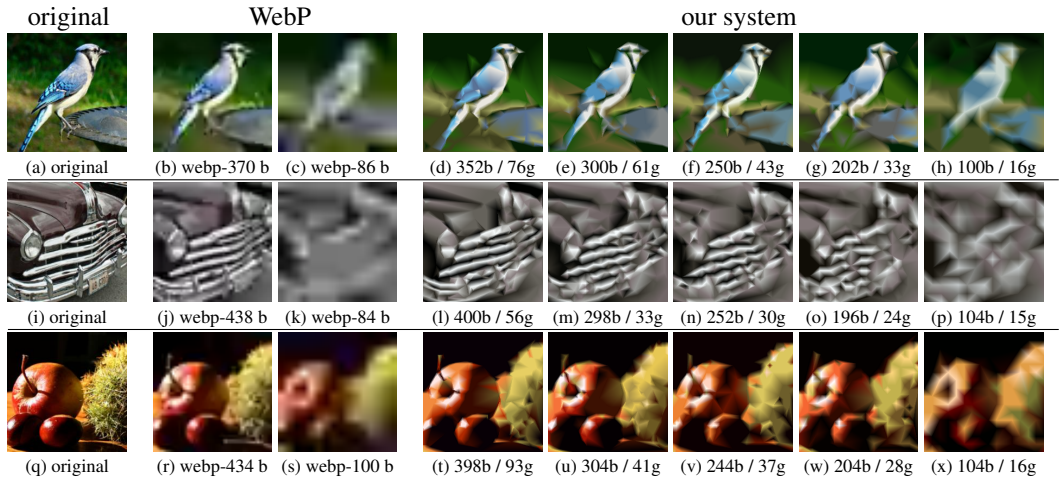}
  \end{minipage}

  \begin{minipage}{5in}
  \end{minipage}
\vskip -0.1in

\end{figure*}

\subsection {Encoding the Triangulation}

Once we have selected a triangulation, we need to losslessly compress
that representation.  In all cases discussed below, we use asymmetric
numerical systems (ANS)~\cite{Duda2015} which is a computationally
efficient method to achieve the compression rates of arithmetic
encoding~\cite{mackay2003}.

We send a ``header'' containing basic parameters such as the grid size,
the number of vertices, and the color table.
The color-table entries
are sorted by frequency of use (starting with the most common).  The
color table is then
coded by giving the number of colors, the color channel
values (quantized to 6 bits per YCoCg channel), and the usage frequencies for each of the
entries. We code the color channel values as a correction from the average
of the previously transmitted values for that channel (e.g., the Y
channel), with the first entry's prediction set to the mid-point gray.
The histogram table for these color differences are sent as part of the
header.

The color-entry frequencies are transmitted using a binomial
distribution model: we use a ``fair selection'' model on the remaining
colors and vertices (i.e., $N = V_{C_r}$ trials, where
$V_{C_r}$ is the number of
remaining vertices not counted by the previous color entries, and 
a success probability of $p = 1/|C_r|$ where $|C_r|$
is the number of remaining color-table entries, $C_r$).
We improve on this probability model by limiting
the considered values of the distribution to be at least
$V_{C_r}/|C_r|$ (based on knowing that the {\em upcoming}
entries have been sorted) and at most the frequency count of the
previous entry (based on knowing that the {\em previous} and {\em current} entries
have been sorted).  Using this approach to color-entry frequency
coding, we save 1\%-3\% of the file size, on average, compared with
using a simple uniform-probability model of the distributions.

We also explore alternative approaches to compressing the vertex
locations and their color assignments.  For the vertex locations, the
simplest alternative is to send one bit (without arithmetic coding)
per grid point to mark whether it is used as a vertex. This takes
$N_g$ bits, where $N_g$ is the number of grid points and corresponds to
using a $0.5$ probability of occupancy.  We can do better 
 by using a fixed, but more accurate, probability of occupancy:
$V_t / N_g$ where $V_t$ is the total number of
vertices (sent in the header).  In our experiments, this
fixed-probability only
saved an average of $1\%$ of the single-bit-per-grid-point approach.  We
also tried using run-length codes (coding the run length of
``unoccupied'' grid points between vertices), however the overhead of
sending the distribution of run-lengths made this approach worse.

The best compression in our tests uses an adaptive-probability
approach to compressing the occupancy map.  In this approach, we
simply update our probability model as we progress through the grid,
so that at each point the model probability is $V_r /
N_r$ where $V_r$ and $N_r$ are the remaining
vertex and grid point counts, respectively.  In our tests, this gave a
savings of $2.25\%$ over the single-bit-per-grid-point approach.

We use a similar approach for coding the color-table index for each vertex.  Since we know the color-entry frequencies, we can use
adaptive models for these indices.
Instead of simply using these remaining-count probabilities,
we can explicitly treat the
color index coding as a chain of Boolean encodings --- each with the
probability indicated by the remaining-count probability for the
corresponding color-table entry but with their order of encoding set
by a spatially adaptive prediction.  This helps even without changing
any of the model probabilities because once we see a ``true'' Boolean
value, we can stop encoding for that vertex and move to the next one.
We determine the order of encoding by sorting the
previously seen (and already transmitted) colors based on their
Manhattan distance
to the current vertex, with ties broken in favor of the more probable color.
Using this ordering results in a $1\%-3\%$ file size reduction 
for the color-table frequencies.  Using the
spatially-adaptive ordering provides an additional $0.66\%$ file-size
reduction.

\section{Experiments}
\label{sec:experiments}

In this section, we present a summary of the experiments
we performed.  The performance is assessed with \emph{PSNR}
(Peak-Signal-to-Noise Ratio) and \emph{SSIM} (Structural Similarity
Index)~\cite{wang2004image}, which is based on the visible structures
in the image, and is considered a perceptual metric.  All of the
results reported are the average of compressing 1,024 images, each
consisting of $221 \times 221$ pixels. The images were randomly selected from
the ImageNet training set~\cite{deng2014imagenet}.
Figure~\ref{mainimages} shows the effects of final byte-size on the
results obtained by the stochastic variant described in the previous
section.  The larger the allowed byte-size, the finer the initial grid
can be.  Our results are shown for grid sizes from $15 \times 15$
(around 100 bytes, compressed) up to $96 \times 96$ (around
400 bytes).

\begin{figure}
  \centering
  \includegraphics[width=\linewidth,height=1.5in]{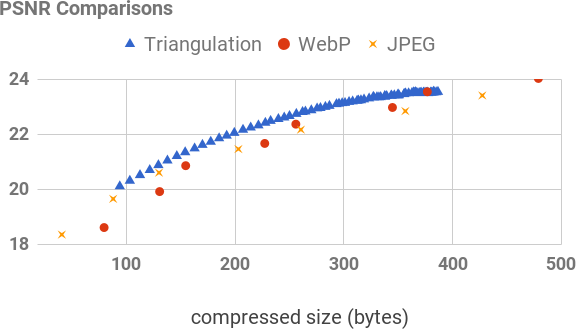}
  \includegraphics[width=\linewidth,height=1.5in]{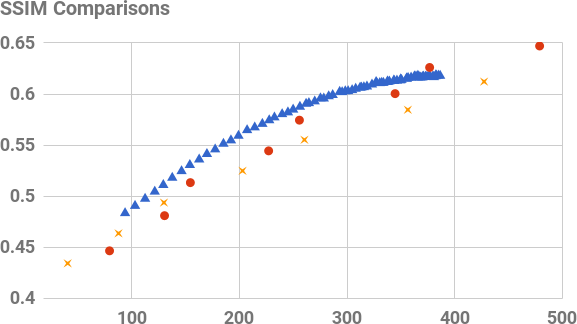}    
  \caption{Measuring quality as a function of bytes. WebP becomes
    competitive at approximately 400 bytes, JPEG is competitive after
    500 (not shown).}
  \label{mainresult}
\vskip -0.2in  
\end{figure}

Quantitative results are shown in Figure~\ref{mainresult}.  When
measured with either metric, PSNR or SSIM, the triangulation approach
outperforms both WebP and JPEG.  As can be seen, JPEG performs quite
poorly in this operating range.  Also note that to be as favorable to
JPEG as possible, we set quality=20, which produced the best
quality/size, and used headerless JPEG encoding.  Adding a header
significantly deteriorates performance. WebP is a close competitor at
quality=10.  At the range of interest (200 bytes) we outperform WebP
in both metrics.  By 400 bytes, WebP and the triangulation approach
perform equally.  Beyond 400 bytes, we again expect WebP to have an
advantage.

\begin{figure}

\centering  
    \includegraphics[width=0.495\linewidth]{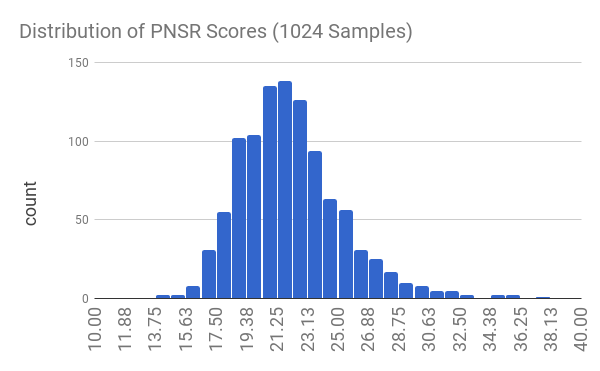}
    \includegraphics[width=0.495\linewidth]{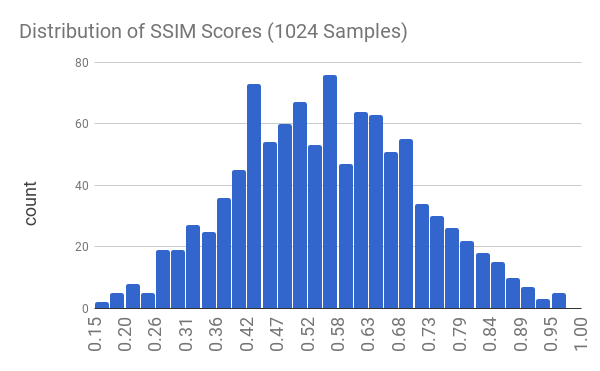}
~\\    
    \footnotesize \rotatebox{90}{~~~~~~$\approx 15$}
    \includegraphics[width=0.151\linewidth]{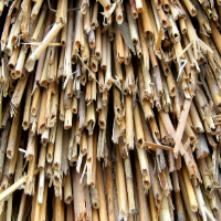}
    \includegraphics[width=0.151\linewidth]{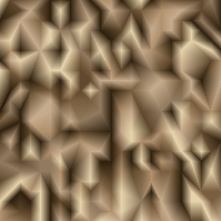}    
    \includegraphics[width=0.151\linewidth]{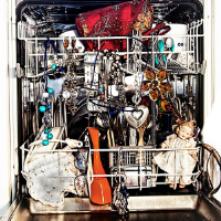}
    \includegraphics[width=0.151\linewidth]{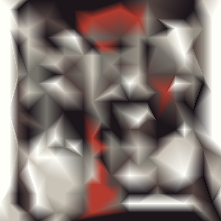}    
    \includegraphics[width=0.151\linewidth]{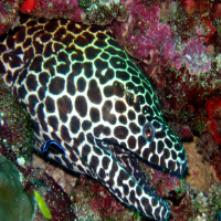}
    \includegraphics[width=0.151\linewidth]{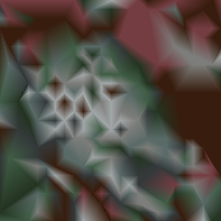}    

    \footnotesize \rotatebox{90}{$psnr\approx 35$}    
    \includegraphics[width=0.151\linewidth]{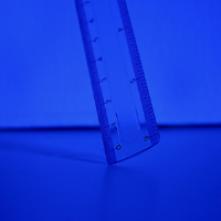}
    \includegraphics[width=0.151\linewidth]{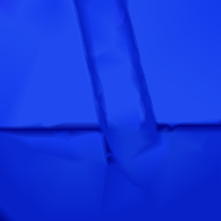}    
    \includegraphics[width=0.151\linewidth]{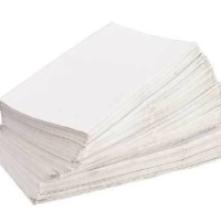}
    \includegraphics[width=0.151\linewidth]{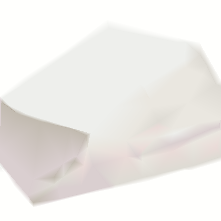}        
    \includegraphics[width=0.151\linewidth]{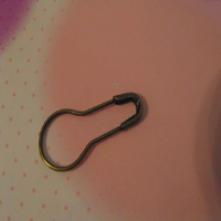}
    \includegraphics[width=0.151\linewidth]{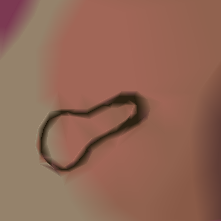}            
    \caption{\small Histogram of Results (at bytes=200).  Top
      Left: PSNR, Top Right: SSIM.  3 Cases with poor (middle row) and
      high PSNR (last row).  Shown alternating: original, compressed. }

    \label {hists}
\vskip -0.1in    
\end{figure}

Despite the promising results in Figure~\ref{mainresult}, we need to
ensure that the average performance is indicative of expected
performance.  In Figure~\ref{hists}, we provide histograms of the PSNR
and SSIM errors on the 1,024 image test set.  In the same figure, we
also look at the best and worst performing examples: intuitively,
the worst cases have large entropy regions
(similar to checkerboard patterns) while the best ones have large
areas of similar colors.

\begin{figure}[b!]
  \vskip -0.1in
  
  \centering
  \includegraphics[width=0.9\linewidth]{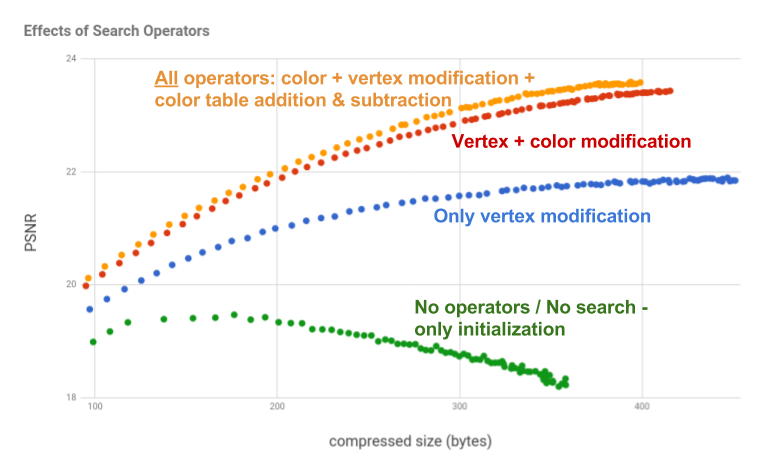}
\vskip -0.1in  
\caption{\small What makes the algorithm (Figure~\ref{stochastic}) work?
  Green: no search, only initialization.
  Blue: using
    only vertex-modification operators (steps~5 (a)-(c)).  Red: adding color
    perturbation (steps~5 (a)-(c) plus (d) and~(g)).  Yellow: all operators
    (steps~5 (a)-(g)).
    }
  
  \label{ablative}

\end{figure}

Finally, we would like to give a better understanding at the best steps
in our pipeline. In the interest of space, we present one graph in
Figure~\ref{ablative} that provides the most insight into the benefits
of the stochastic approach over the baseline.  We repeat all of our
experiments using only a subset of the operators described in
Figure~\ref{stochastic} Step 5. The biggest improvement is from
vertex modfication and, next, from color-entry perturbation (PSNR shown, same for SSIM).
The gains seen with color modification suggest that pre-computing the color table based
on the color clusters in the thumbnail
is not adequate
for use in a triangulated approximation.

\section {Conclusions \& Future Work}

We have presented an approach to compressing images to extremely small
byte sizes.  In the operating range of interest, 200-400 bytes,
standard JPEG operates poorly.  With tiny thumbnails under 400 bytes,
we surpass the latest deployed version WebP, version 1.0.0, in both
PSNR and SSIM.  Only after $\sim$400 bytes, does WebP do better.
Further, our representation is scale-free: the triangles can be easily
scaled and simply rendered with low-level primitives.

There are three immediate avenues for future work.  The first is the large-scale human evaluation of the images to verify the SSIM and PSNR improvements.
Second, the stochastic search process can
be computationally expensive; domain-specific heuristics to narrow the search space may help.
Third, we have
found that adding synthetic noise in post-processing can enhance
the perceived quality without always improving
the quality/size score. Further research on a perceptual metric dedicated
to thumbnails is needed.

\bibliography{triangle}

\bibliographystyle{IEEEbib}

\begin{figure*}[t]
\section*{Supplemental Materials}

The below are seven example images, which we have taken from our evaluation
set.  The subsequent pages show them encoded with different sized
grids, across our full range of bitrates.  Below each image, we
list PSNR, SSIM, and file size.  We use file size, instead of BPP,
since our triangle representation does not limit the reconstruction
to any fixed size.  The reconstructions are shown at
$221 \times 221$, but larger-sized reconstructions will retain the sharpness of
these reconstructions, since the upsampling conversion is done
before the interpolation, used to fill in the Delaney triangles.
The ridges that are seen in the $221 \times 221$ reconstructions will remain
as sharp, even at larger image sizes, due to the vertex-based representation.
\end{figure*}

\begin{figure*}[h]
\begin{tabular}{cccc}
  \includegraphics[width=1.51in]{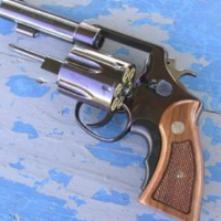} &
  \includegraphics[width=1.51in]{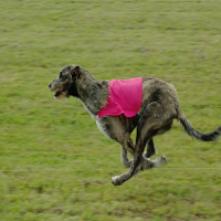} &
  \includegraphics[width=1.51in]{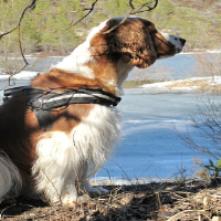} &
  \includegraphics[width=1.51in]{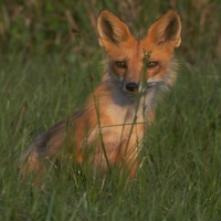} \\
  \includegraphics[width=1.51in]{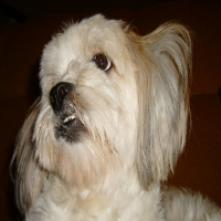} &
  \includegraphics[width=1.51in]{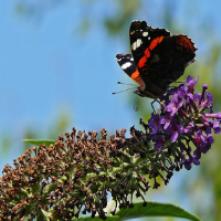} &
  \includegraphics[width=1.51in]{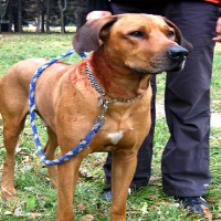} \\
\end{tabular}
  \caption*{Original target images}
\end{figure*}

\begin{figure*}
\begin{tabular}{p{2.0in}p{2.0in}p{2.0in}}
  \includegraphics[width=2.0in]{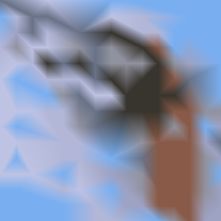} \footnotesize PSNR: 20.18; SSIM: 0.515; bytes: 98.0 &
  \includegraphics[width=2.0in]{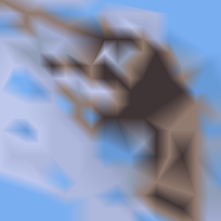} \footnotesize PSNR: 20.43; SSIM: 0.531; bytes: 128.0 &
  \includegraphics[width=2.0in]{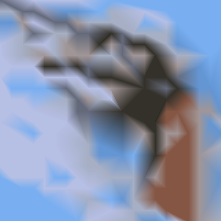} \footnotesize PSNR: 21.15; SSIM: 0.550; bytes: 152.0 \\
  \includegraphics[width=2.0in]{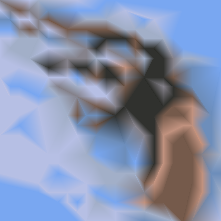} \footnotesize PSNR: 21.43; SSIM: 0.565; bytes:  176.0 &
  \includegraphics[width=2.0in]{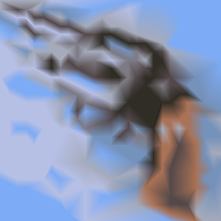} \footnotesize PSNR: 21.76; SSIM: 0.578; bytes: 208.0 &
  \includegraphics[width=2.0in]{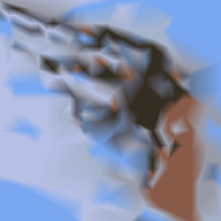} \footnotesize PSNR: 22.06; SSIM: 0.590; bytes: 236.0 \\
  \includegraphics[width=2.0in]{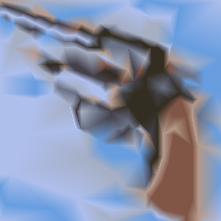} \footnotesize PSNR: 22.48; SSIM: 0.609; bytes: 256.0 &
  \includegraphics[width=2.0in]{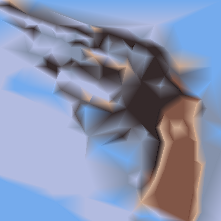} \footnotesize PSNR: 22.63; SSIM: 0.617; bytes: 278.0 &
  \includegraphics[width=2.0in]{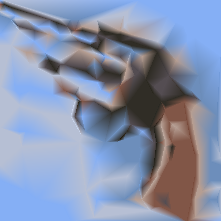} \footnotesize PSNR: 22.66; SSIM: 0.622; bytes:  298.0 \\
  \includegraphics[width=2.0in]{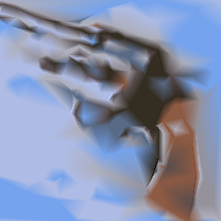} \footnotesize PSNR: 22.73; SSIM: 0.624; bytes:  314.0 &
  \includegraphics[width=2.0in]{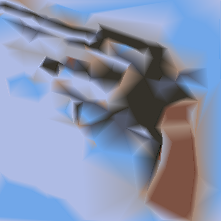} \footnotesize PSNR: 22.86; SSIM: 0.635; bytes:  344.0 &
  \includegraphics[width=2.0in]{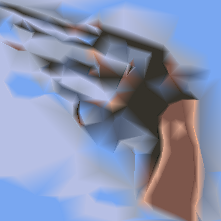} \footnotesize PSNR: 23.20; SSIM: 0.643; bytes:  370.0
\end{tabular}
\end{figure*}

\begin{figure*}
\begin{tabular}{p{2.0in}p{2.0in}p{2.0in}}
  \includegraphics[width=2.0in]{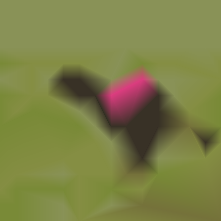} \footnotesize PSNR: 24.31; SSIM: 0.578; bytes:  70.0 &
  \includegraphics[width=2.0in]{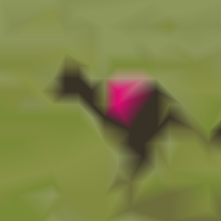} \footnotesize PSNR: 24.80; SSIM: 0.586; bytes:  120.0 &
  \includegraphics[width=2.0in]{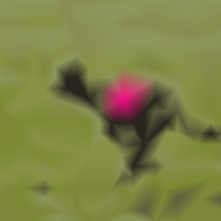} \footnotesize PSNR: 25.14; SSIM: 0.594; bytes:  140.0 \\
  \includegraphics[width=2.0in]{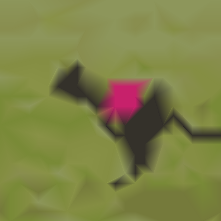} \footnotesize PSNR: 25.28; SSIM: 0.604; bytes:  168.0 &
  \includegraphics[width=2.0in]{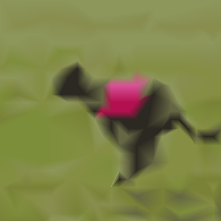} \footnotesize PSNR: 26.11; SSIM: 0.617; bytes:  192.0 &
  \includegraphics[width=2.0in]{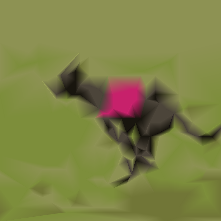} \footnotesize PSNR: 26.52; SSIM: 0.635; bytes:  222.0 \\
  \includegraphics[width=2.0in]{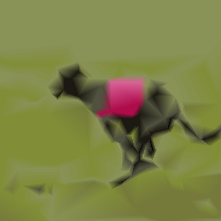} \footnotesize PSNR: 27.24; SSIM: 0.648; bytes:  246.0 &
  \includegraphics[width=2.0in]{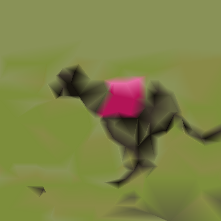} \footnotesize PSNR: 26.97; SSIM: 0.643; bytes:  264.0 &
  \includegraphics[width=2.0in]{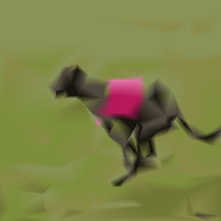} \footnotesize PSNR: 27.38; SSIM: 0.654; bytes:  288.0 \\
  \includegraphics[width=2.0in]{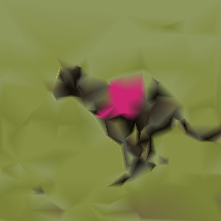} \footnotesize PSNR: 27.03; SSIM: 0.649; bytes:  312.0 &
  \includegraphics[width=2.0in]{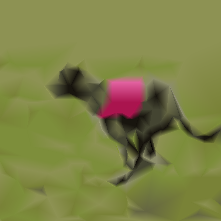} \footnotesize PSNR: 27.57; SSIM: 0.660; bytes:  340.0 &
  \includegraphics[width=2.0in]{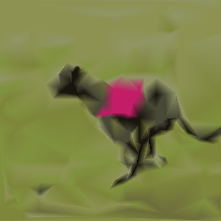} \footnotesize PSNR: 27.20; SSIM: 0.659; bytes:  352.0
\end{tabular}
\end{figure*}     

\begin{figure*}
\begin{tabular}{p{2.0in}p{2.0in}p{2.0in}}
  \includegraphics[width=2.0in]{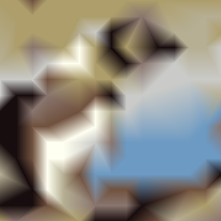} \footnotesize PSNR: 18.61; SSIM: 0.410; bytes:  96.0 &
  \includegraphics[width=2.0in]{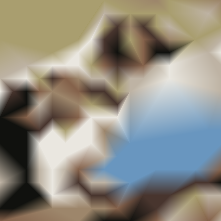} \footnotesize PSNR: 18.97; SSIM: 0.422; bytes:  116.0 &
  \includegraphics[width=2.0in]{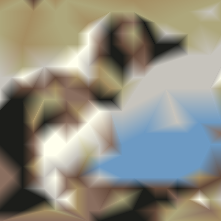} \footnotesize PSNR: 19.64; SSIM: 0.453; bytes:  154.0 \\
  \includegraphics[width=2.0in]{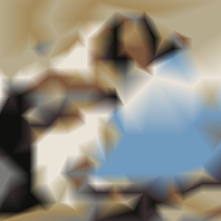} \footnotesize PSNR: 19.72; SSIM: 0.454; bytes:  178.0 &
  \includegraphics[width=2.0in]{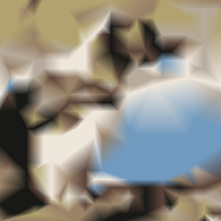} \footnotesize PSNR: 20.08; SSIM: 0.483; bytes:  194.0 &
  \includegraphics[width=2.0in]{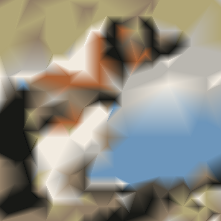} \footnotesize PSNR: 20.36; SSIM: 0.479; bytes:  230.0 \\
  \includegraphics[width=2.0in]{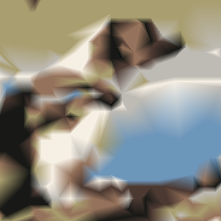} \footnotesize PSNR: 20.72; SSIM: 0.508; bytes:  268.0 &
  \includegraphics[width=2.0in]{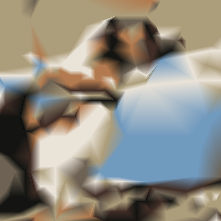} \footnotesize PSNR: 20.80; SSIM: 0.517; bytes:  286.0 &
  \includegraphics[width=2.0in]{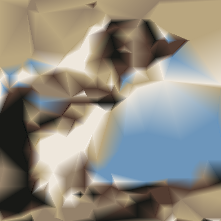} \footnotesize PSNR: 20.72; SSIM: 0.514; bytes:  302.0 \\
  \includegraphics[width=2.0in]{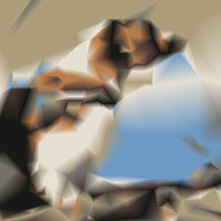} \footnotesize PSNR: 20.91; SSIM: 0.520; bytes:  328.0 &
  \includegraphics[width=2.0in]{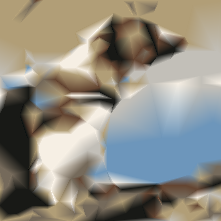} \footnotesize PSNR: 21.15; SSIM: 0.532; bytes:  344.0 &
  \includegraphics[width=2.0in]{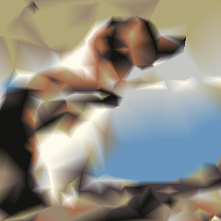} \footnotesize PSNR: 21.39; SSIM: 0.532; bytes:  374.0
\end{tabular}
\end{figure*}     

\begin{figure*}
\begin{tabular}{p{2.0in}p{2.0in}p{2.0in}}
  \includegraphics[width=2.0in]{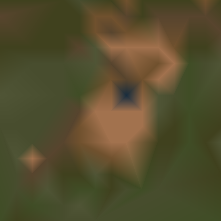} \footnotesize PSNR: 26.22; SSIM: 0.625; bytes:  86.0 &
  \includegraphics[width=2.0in]{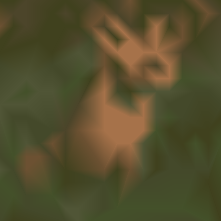} \footnotesize PSNR: 26.59; SSIM: 0.635; bytes:  118.0 &
  \includegraphics[width=2.0in]{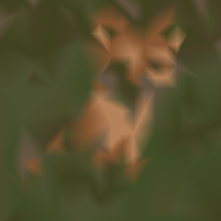} \footnotesize PSNR: 27.36; SSIM: 0.648; bytes:  140.0 \\
  \includegraphics[width=2.0in]{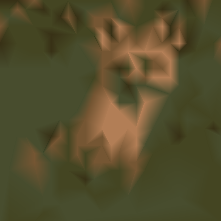} \footnotesize PSNR: 27.48; SSIM: 0.662; bytes:  164.0 &
  \includegraphics[width=2.0in]{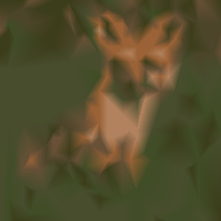} \footnotesize PSNR: 27.91; SSIM: 0.666; bytes:  204.0 &
  \includegraphics[width=2.0in]{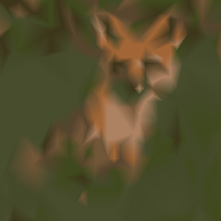} \footnotesize PSNR: 28.27; SSIM: 0.675; bytes:  248.0 \\
  \includegraphics[width=2.0in]{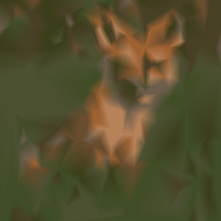} \footnotesize PSNR: 28.50; SSIM: 0.682; bytes:  282.0 &
  \includegraphics[width=2.0in]{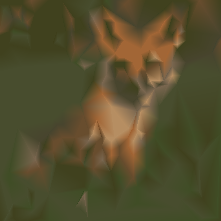} \footnotesize PSNR: 28.78; SSIM: 0.689; bytes:  308.0 &
  \includegraphics[width=2.0in]{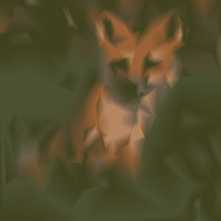} \footnotesize PSNR: 28.79; SSIM: 0.691; bytes:  336.0 \\
  \includegraphics[width=2.0in]{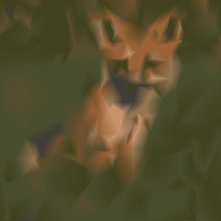} \footnotesize PSNR: 28.77; SSIM: 0.689; bytes:  354.0 &
  \includegraphics[width=2.0in]{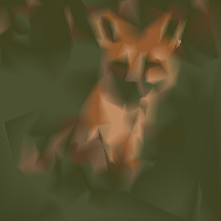} \footnotesize PSNR: 28.76; SSIM: 0.693; bytes:  370.0 &
  \includegraphics[width=2.0in]{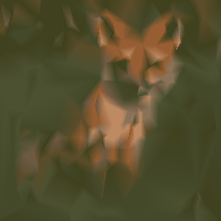} \footnotesize PSNR: 29.04; SSIM: 0.701; bytes:  382.0 \\
\end{tabular}
\end{figure*}     

\begin{figure*}
\begin{tabular}{p{2.0in}p{2.0in}p{2.0in}}
  \includegraphics[width=2.0in]{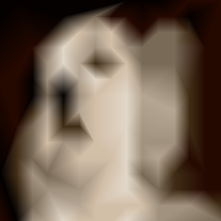} \footnotesize PSNR: 23.50; SSIM: 0.623; bytes:  94.0 &
  \includegraphics[width=2.0in]{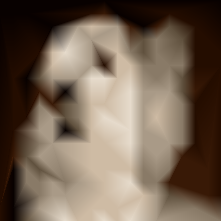} \footnotesize PSNR: 24.15; SSIM: 0.653; bytes:  116.0 &
  \includegraphics[width=2.0in]{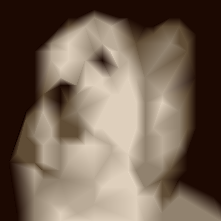} \footnotesize PSNR: 24.95; SSIM: 0.670; bytes:  134.0 \\
  \includegraphics[width=2.0in]{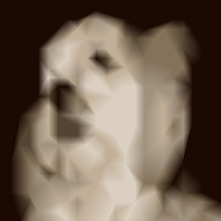} \footnotesize PSNR: 25.26; SSIM: 0.683; bytes:  146.0 &
  \includegraphics[width=2.0in]{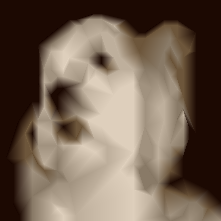} \footnotesize PSNR: 26.18; SSIM: 0.709; bytes:  198.0 &
  \includegraphics[width=2.0in]{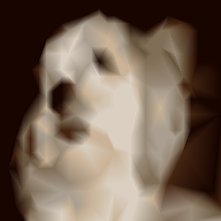} \footnotesize PSNR: 26.35; SSIM: 0.707; bytes:  208.0 \\
  \includegraphics[width=2.0in]{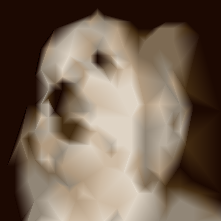} \footnotesize PSNR: 26.38; SSIM: 0.702; bytes:  216.0 &
  \includegraphics[width=2.0in]{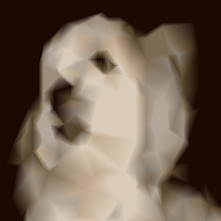} \footnotesize PSNR: 26.47; SSIM: 0.712; bytes:  238.0 &
  \includegraphics[width=2.0in]{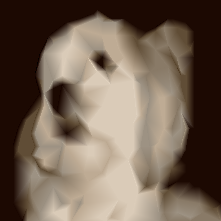} \footnotesize PSNR: 26.59; SSIM: 0.721; bytes:  258.0 \\
  \includegraphics[width=2.0in]{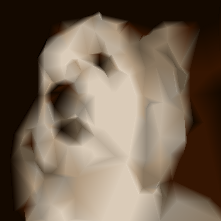} \footnotesize PSNR: 27.13; SSIM: 0.735; bytes:  288.0 &
  \includegraphics[width=2.0in]{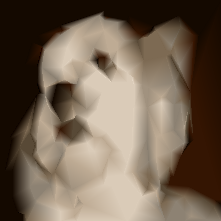} \footnotesize PSNR: 27.60; SSIM: 0.749; bytes:  332.0 &
  \includegraphics[width=2.0in]{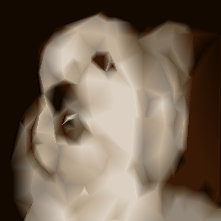} \footnotesize PSNR: 27.78; SSIM: 0.748; bytes:  342.0
\end{tabular}
\end{figure*}     

\begin{figure*}
\begin{tabular}{p{2.0in}p{2.0in}p{2.0in}}
  \includegraphics[width=2.0in]{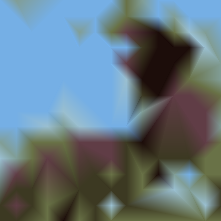} \footnotesize PSNR: 18.42; SSIM: 0.494; bytes:  90.0 &
  \includegraphics[width=2.0in]{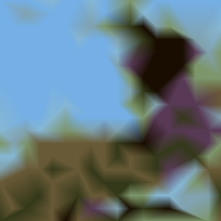} \footnotesize PSNR: 18.74; SSIM: 0.503; bytes:  124.0 &
  \includegraphics[width=2.0in]{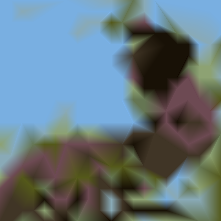} \footnotesize PSNR: 19.08; SSIM: 0.524; bytes:  152.0 \\
  \includegraphics[width=2.0in]{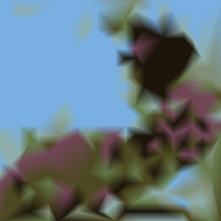} \footnotesize PSNR: 19.19; SSIM: 0.531; bytes:  170.0 &
  \includegraphics[width=2.0in]{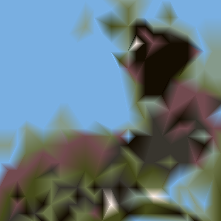} \footnotesize PSNR: 19.56; SSIM: 0.546; bytes:  204.0 &
  \includegraphics[width=2.0in]{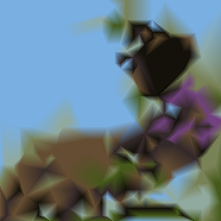} \footnotesize PSNR: 19.90; SSIM: 0.564; bytes:  252.0 \\
  \includegraphics[width=2.0in]{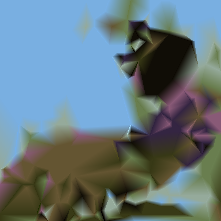} \footnotesize PSNR: 20.09; SSIM: 0.574; bytes:  312.0 &
  \includegraphics[width=2.0in]{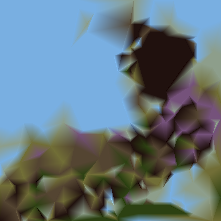} \footnotesize PSNR: 20.18; SSIM: 0.586; bytes:  336.0 &
  \includegraphics[width=2.0in]{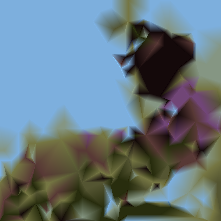} \footnotesize PSNR: 20.27; SSIM: 0.593; bytes:  352.0 \\
  \includegraphics[width=2.0in]{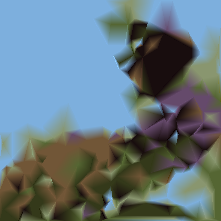} \footnotesize PSNR: 20.30; SSIM: 0.592; bytes:  378.0 &
  \includegraphics[width=2.0in]{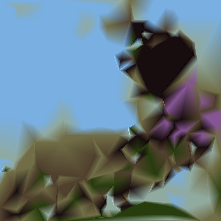} \footnotesize PSNR: 20.39; SSIM: 0.600; bytes:  396.0 &
  \includegraphics[width=2.0in]{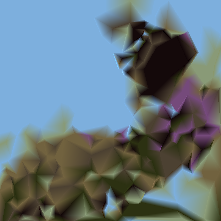} \footnotesize PSNR: 20.43; SSIM: 0.600; bytes:  416.0
\end{tabular}
\end{figure*}     

\begin{figure*}
\begin{tabular}{p{2.0in}p{2.0in}p{2.0in}}
  \includegraphics[width=2.0in]{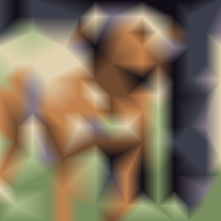} \footnotesize PSNR: 17.02; SSIM: 0.328; bytes:  102.0 &
  \includegraphics[width=2.0in]{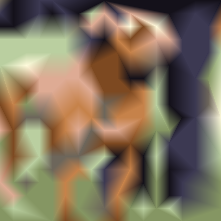} \footnotesize PSNR: 17.39; SSIM: 0.336; bytes:  132.0 &
  \includegraphics[width=2.0in]{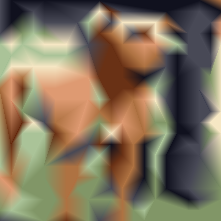} \footnotesize PSNR: 17.77; SSIM: 0.360; bytes:  154.0 \\
  \includegraphics[width=2.0in]{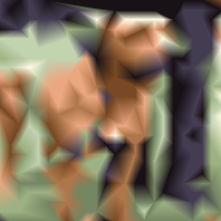} \footnotesize PSNR: 18.28; SSIM: 0.387; bytes:  200.0 &
  \includegraphics[width=2.0in]{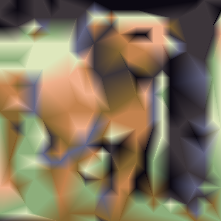} \footnotesize PSNR: 18.47; SSIM: 0.395; bytes:  214.0 &
  \includegraphics[width=2.0in]{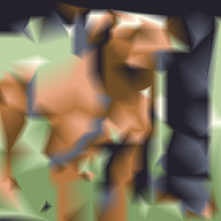} \footnotesize PSNR: 18.78; SSIM: 0.405; bytes:  242.0 \\
  \includegraphics[width=2.0in]{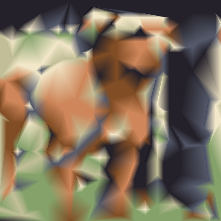} \footnotesize PSNR: 18.97; SSIM: 0.424; bytes:  284.0 &
  \includegraphics[width=2.0in]{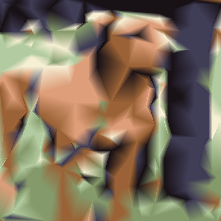} \footnotesize PSNR: 19.45; SSIM: 0.452; bytes:  340.0 &
  \includegraphics[width=2.0in]{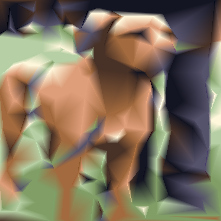} \footnotesize PSNR: 19.54; SSIM: 0.456; bytes:  388.0 \\
  \includegraphics[width=2.0in]{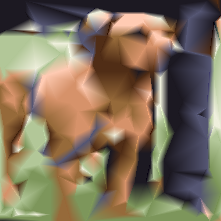} \footnotesize PSNR: 19.83; SSIM: 0.470; bytes:  398.0 &
  \includegraphics[width=2.0in]{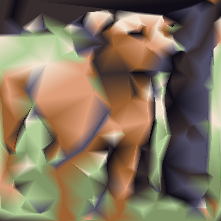} \footnotesize PSNR: 19.84; SSIM: 0.470; bytes:  456.0 &
  \includegraphics[width=2.0in]{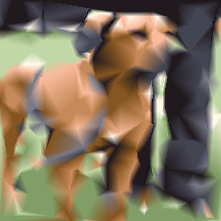} \footnotesize PSNR: 20.02; SSIM: 0.484; bytes:  472.0
\end{tabular}
\end{figure*}

\end{document}